\documentclass[10pt, a4paper]{article}

\usepackage[final]{lrec2026} 
\usepackage{listings}
\usepackage{longtable}
\usepackage{hhline}
\usepackage{latexsym}
\usepackage{amsmath}
\usepackage{amssymb}
\usepackage{amsfonts}
\usepackage{bbm}
\usepackage{multirow}
\usepackage{adjustbox}
\lstdefinestyle{nicecode}{
  basicstyle=\ttfamily\footnotesize,
  columns=fullflexible,
  keepspaces=true,
  breaklines=true,
  breakatwhitespace=true,
  showstringspaces=false,
  frame=single,
  rulecolor=\color{black!15},
  frameround=tttt,
  tabsize=2,
  upquote=true,
  keywordstyle=\bfseries\color{darkblue},
  commentstyle=\itshape\color{black!55},
  stringstyle=\color{black!70},
}

\title{BLooP: Zero-Shot Abstractive Summarization using\\Large Language Models with Bigram Lookahead Promotion}

\name{Varun Iyer \mbox{ }\mbox{ }\mbox{ } \mbox{ }\mbox{ }\mbox{ }\mbox{ }\mbox{ }\mbox{ } Cornelia Caragea} 

\address{University of Illinois Chicago \\
viyer9@uic.edu  \mbox{ }\mbox{ }\mbox{ }\mbox{ }\mbox{ }\mbox{ } cornelia@uic.edu
         \\}

\abstract{
Abstractive summarization requires models to generate summaries that convey information in the source document. While large language models can generate summaries without fine-tuning, they often miss key details and include extraneous information. We propose BLooP (Bigram Lookahead Promotion), a simple training-free decoding intervention that encourages large language models (LLMs) to generate tokens that form bigrams from the source document. BLooP operates through a hash table lookup at each decoding step, requiring no training, fine-tuning, or model modification. We demonstrate improvements in ROUGE and BARTScore for Llama-3.1-8B-Instruct, Mistral-Nemo-Instruct-2407, and Gemma-2-9b-it on CNN/DM, CCSum, Multi-News, and SciTLDR. Human evaluation shows that BLooP significantly improves faithfulness without reducing readability. We make the code available here: \url{https://github.com/varuniyer/BLooP}.
 \\ \newline  }

\begin{document}

\maketitleabstract

\section{Introduction}

Abstractive summarization requires models to generate concise summaries that capture the key points of source documents while maintaining both readability and faithfulness to the original content \cite{rush-etal-2015-neural, nallapati, pegasus, zhou-selective, fact-cons}. Modern approaches typically employ transformer-based architectures \cite{attn} and frame the task as a conditional sequence-to-sequence problem \cite{bart, pegasus, prom}. Pre-trained language models trained on large-scale corpora \cite{bart, t5, pegasus, palm, prophetnet, prom} have recently emerged as the dominant paradigm in this task.

Despite significant progress, generating faithful summaries remains challenging. While copy mechanisms were introduced to bridge extraction and generation by allowing models to select tokens directly from source documents \cite{ptrgen, coconet}, even these enhanced models struggle with faithfulness due to exposure bias \cite{brio}, insufficient document comprehension \cite{bass, gsum}, and overconfident predictions \cite{faith-abs}. Traditional copy mechanisms operate by learning to interpolate between a copying distribution and the language model's generation distribution, requiring substantial training data to function effectively.

This data dependency presents a critical bottleneck. Conventional abstractive summarization models require thousands of manually annotated article-summary pairs for training \citeplanguageresource{cnndm} \cite{nallapati, narayan}. While approaches like PEGASUS \cite{pegasus} attempt to leverage unlabeled data through gap sentence generation (GSG), the fundamental need for domain-specific fine-tuning persists. Recent large language models (LLMs) have demonstrated impressive zero-shot summarization capabilities \cite{llm-benchmarking}, yet they are still predominantly fine-tuned on target datasets for optimal performance \cite{news-review, factual, xscitldr, difftopk}. This requirement for extensive annotated data becomes prohibitive when deploying summarization systems to new domains or specialized contexts.

We propose \textbf{BLooP} (\textbf{B}i-gram \textbf{Loo}kahead \textbf{P}romotion), a training-free copy mechanism that addresses this limitation by enabling LLMs to generate more faithful summaries without any dataset-specific fine-tuning. Unlike pointer-generator networks that require learning attention-based copying distributions, BLooP operates purely during decoding by proactively biasing token probabilities toward source document bigrams at each generation step. When the previously generated token concatenated with a candidate next token matches a bigram in the source, BLooP increases that candidate's probability. Our BLooP approach requires no learned parameters, operates entirely in the vocabulary space, and works with any decoder-only LLM through a single tunable hyperparameter ($\alpha$).

The key insight behind BLooP is that by maintaining a bigram cache and proactively boosting tokens that extend the current context into the source-document bigrams, we can ground the summary more closely to factual content while preserving the model's natural generation capabilities. This simple yet effective mechanism reduces hallucination and improves faithfulness without sacrificing fluency, as demonstrated in our experiments.

Our contributions are as follows:

\begin{enumerate}
  \item We introduce BLooP, a training-free decoding method that adjusts token probabilities based on bigram matches with the input document, requiring only a single hyperparameter.
  \item We demonstrate consistent improvements in ROUGE and BARTScore metrics across multiple datasets using Llama-3.1-8B-Instruct \cite{llama3}, Gemma-2-9b-it \cite{gemma2}, and Mistral-Nemo-Instruct-2407 (12B)\footnote{\url{https://mistral.ai/en/news/mistral-nemo}}.
  \item We conduct human evaluation showing that BLooP significantly improves summary faithfulness while maintaining informativeness and readability.
  \item We provide a detailed analysis of the bigram cache mechanism and its effectiveness across different summarization contexts.
\end{enumerate}

\section{Related Work}

Abstractive summarization has undergone significant evolution with the advent of neural sequence-to-sequence models. Pointer-generator networks \cite{ptrgen} addressed early challenges in abstractive summarization by enabling models to both generate novel text and copy words from the source document.

The pointer-generator mechanism uses hidden representations of article tokens to compute an attention distribution over these tokens at each generation timestep. Additionally, an attention-weighted average of the encoder hidden states is passed into an MLP to compute a copy probability. This copy probability is used to linearly interpolate the language model's next token distribution and the attention distribution over input article tokens. Finally, the next token is sampled from this interpolated distribution. This approach requires significant computational overhead to train the model end-to-end, including both the attention mechanism and the MLP.

Since the emergence of the pointer-generator mechanism, researchers have explored advanced techniques to enhance summary quality using encoder-decoder transformer architectures such as BART \cite{bart}. For example, PEGASUS \cite{pegasus} introduced pre-training on a gap sentence generation task to improve summarization performance. Its architecture uses an attention-weighted average of encoder hidden states to generate summaries, enabling the model to copy and generate text effectively. To encourage the model to plan for future tokens, ProphetNet \cite{prophetnet} proposes an $n$-stream self-attention mechanism as an extension of the two-stream self-attention introduced by XLNet \cite{xlnet}. It is trained to simultaneously predict $n$ tokens at each generation step in addition to the traditional autoregressive language modeling objective. 

Later, PROM \cite{prom} was introduced as an improvement over PEGASUS. PROM uses a copying mechanism that explicitly identifies $n$-grams suitable for copying, demonstrating significant improvements in both fine-tuned and zero-shot summarization settings. Their model is pre-trained on the gap sentence generation (GSG) task introduced by PEGASUS (\cite{pegasus}). Their work highlights the continued importance of copying spans from the input document to improve summary quality. It also demonstrates superior performance to decoder-only large language models (LLMs) such as Llama 2 \cite{llama2} in the zero-shot setting. While PROM demonstrates the continued importance of copying mechanisms, it requires specialized pre-training on gap sentence generation tasks and additional model parameters, making it less suitable for zero-shot applications with existing LLMs.

The faithfulness of zero-shot summarization remains challenging even for well-trained models. Past work found that GPT-3.5 \cite{gpt3} and FLAN-T5 \cite{flan} introduce more extrinsic errors (hallucinations) in news summaries than in legal and biomedical document summaries even though news articles have significantly greater representation in LLM training data \cite{evaluating}. This work highlights that faithfulness challenges persist across domains, motivating training-free approaches that can improve source grounding without domain-specific fine-tuning.

Recently, \citet{largepig} introduced LargePiG, an approach that uses decoder-only LLMs' attention weights in a pointer-generator mechanism. However, this requires a computationally expensive multi-step process: (1) Extracting and normalizing attention weights from specific layers; (2) Constructing pointer attention distributions that match vocabulary distributions; and (3) Computing copy probabilities through early-exit mechanisms that compare distributions across multiple transformer layers. This approach requires accessing internal model states and substantial computational overhead. In contrast, BLooP operates purely at the vocabulary level with a simple hash table lookup at each timestep. LargePiG confirms that extracting usable attention weights from modern LLMs for copying mechanisms is non-trivial and computationally expensive, further highlighting BLooP's simplicity and efficiency advantages. Moreover, efficient attention implementations such as PagedAttention \cite{vllm} do not store the entire attention matrix in memory. Thus, full attention matrices are not readily available when using vLLM and other optimized LLM implementations. Incompatibility with these efficient attention implementations significantly lowers inference speed.

Further work has introduced a lookahead heuristic that scores the faithfulness of potential future summaries during generation \cite{lookahead}. However, this approach incurs a quadratic number of generation steps relative to the summary length, as it requires generating a full summary for candidate tokens at every timestep. The $O(n^2)$ complexity is computationally infeasible for multi-billion parameter models. 
This work iteratively distills a teacher model using expensive lookahead decoding into an identical student that uses greedy decoding. This is guided using a learned composite metric via linear regression on human faithfulness judgments. The composite metric is a linear model trained to predict human faithfulness judgments using some of the same automatic metrics employed in their final evaluation. Thus, this approach is distinct from purely training-free interventions.

Another line of work modifies token generation probabilities during decoding to improve output quality. For example, contrastive search \cite{contrast} penalizes tokens that are too similar to previous context while maintaining model confidence. Coverage mechanisms \cite{coverage, ptrgen} track attention distributions to ensure comprehensive source coverage during generation. Unlike these approaches, which operate on generic linguistic patterns or require additional model components, BLooP specifically targets content fidelity by encouraging tokens that form valid bigrams from the source document.

Our work aims to enhance the quality of LLM-generated summaries without additional training or slower inference. To this end, we introduce a bigram lookahead promotion that provides a simple yet effective mechanism to improve summary quality. By proactively managing token generation based on input document bigrams, we offer a novel, efficient approach to enhancing the quality of generated summaries as measured by ROUGE \cite{rouge} and BARTScore \cite{bartscore}.

\section{Methodology}
\label{method}

\subsection{Task Formulation}
In abstractive summarization, the input is a document $D$. The goal is to generate a summary $S$ that captures the key information in $D$. Let the sentences in $D$ be $D = \{C_1, C_2, \ldots, C_n\}$, where $C_i=\{d_{i,1}, d_{i,2}, \ldots, d_{i,m_i}\}$ is a sequence of tokens belonging to the $i$-th sentence. Given $D$, the task is to predict a summary $S = \{s_1, s_2, \ldots, s_m\}$. With large language models, this is done sequentially. The model's prediction at time-step $t$ is $s_t \in V$. At time-step $t$, the model predicts logits $L \in \mathbb{R}^{\mid V \mid}$, where $V$ is the model's vocabulary. The model's prediction at time-step $t \in \mathbb{Z}^+$ is $s_t = \arg\max_{v \in V} L_v$. We used beam search to generate better summaries by considering the top $k$ candidates at each time-step.

\subsection{Proposed Approach: BLooP}
To improve the quality of generated summaries, we apply a bigram lookahead promotion (BLooP) after the first generation time-step. Our goal is to proactively signal intra-sentence bigram continuations that match source document bigrams. To this end, we first define the set of intra-sentence bigrams in $D$ as follows:

\begin{equation}
  \label{eq:bigram}
  B = \bigcup_{C_i \in D} \{(d_{i, j}, d_{i, j+1}) \mid
  d_{i, j} \in C_i\}
\end{equation}

$L$ is promoted by adding some positive $\alpha$ if the bigram $(s_{t-1}, v) \in B$:

\begin{equation}
  \label{eq:promotion}
  L_v' = L_v + \alpha \cdot \mathbbm{1}[(s_{t-1}, v) \in B], \quad \forall v \in V
\end{equation}

The promotion is applied to tokens that, when concatenated with the previously generated token, result in a bigram from the input document. Bigrams represent the optimal granularity for copying: uni-grams would indiscriminately boost all source tokens regardless of context, while higher-order $n$-grams become exponentially sparse and rarely match generated sequences. This ensures BLooP encourages contextually appropriate token selection from the source.

The token predicted at time-step $t$ before promoting $L$ is $s_t = \arg\max_{v \in V} L_v$. Let $E$ be the set of tokens that mark the end of $S$ (denoting a stop-string that halts the generation process). In our experiments, $E$ contains all tokens in $V$ that contain a newline character. If $s_t \in E$, the promotion in Equation~\ref{eq:promotion} is not applied. This prevents BLooP from artificially extending summaries beyond their natural endpoint by promoting end-of-summary tokens. 

After the promotion is applied, $s_t' = \arg\max_{v \in V} L_v'$ is the token predicted at time-step $t$. This process is repeated until $s_t \in E$. During a beam search, $L'$ is used to score the beam candidates at each time-step. In this context, we define a bigram cache as a data structure that efficiently stores $B$ and supports querying $s_{t-1}$ to return

\begin{equation}
  \label{eq:cache}
  B_{s_{t-1}} = \{v \mid (s_{t-1}, v) \in B\}
\end{equation}

\noindent in $O(1)$ time. This is implemented as a hash table mapping each LLM's token ids to a de-duplicated list of all token ids that follow it in the input document. For example, if the input document was "This dog’s certainly not setting a good example" and each word was a token, $B_{\text{good}}$ would equal $\{\text{example}\}$ because "example" immediately follows "good" in the document. A cache hit occurs when $B_{s_{t-1}} \neq \emptyset$.

Intuitively, this approach is used to encourage tokens that result in the LLM generating bigrams that occur in the input document. This is achieved using $\alpha$ in Equation \ref{eq:promotion}. The same constant promotion is applied to all tokens in $B_{s_{t-1}}$ to ensure that the following equality holds:
\begin{equation}
    \label{equality}
    {\arg\max}_{v \in B_{s_{t-1}}} L_v' = {\arg\max}_{v \in B_{s_{t-1}}} L_v
\end{equation}
This property is desirable because the promotion is only meant to shift the LLM's predictions away or towards tokens in $B_{s_{t-1}}$. As such, the promotion deliberately avoids altering the LLM's next token distribution of tokens in $B_{s_{t-1}}$.

\section{Experimental Setup}

\subsection{Datasets}
We evaluate the effectiveness of BLooP with LLMs on the test splits of the following summarization datasets:

\paragraph{CNN/DM} \citeplanguageresource{cnndm}: 11,490 news articles sourced from CNN and Daily Mail paired with summaries written by the news providers.
\vspace{-2mm}
\paragraph{Multi-News} \citeplanguageresource{multinews}: 5,622 groups of news articles from various sources paired with summaries written by professional editors on the site newser.com.
\vspace{-2mm}
\paragraph{CCSum} \cite{ccsum}: 10,000 news articles from the Common Crawl News corpus \footnote{\url{https://commoncrawl.org/blog/news-dataset-available}} paired with the first sentences of similar articles from different sources. The authors introduce a sophisticated extraction + filtering pipeline to significantly reduce errors in CCSum's reference summaries (even compared to CNN/DM and Multi-News, which use human-written summaries). This dataset's articles are not openly available due to copyright restrictions. Nonetheless, we were granted access to CCSum from its authors. We use this dataset to more accurately assess BLooP's news summarization performance.
\vspace{-2mm}
\paragraph{SciTLDR} \citeplanguageresource{scitldr}: 618 computer science papers paired with expert-annotated summaries written to align with peer reviews from OpenReview \cite{openreview}. SciTLDR summaries can be generated from the abstract alone, the abstract + introduction + conclusion (AIC), and the full text. We used all three versions in our experiments. 

\subsection{Baselines and Evaluation Metrics}
For news summarization, we compare the zero-shot performance of LLMs with and without BLooP to the following baselines:

\vspace{-2mm}
\paragraph{PEGASUS} \cite{pegasus}: An encoder-decoder transformer model pre-trained with the GSG objective to mimic the real abstractive summarization task.
\vspace{-2mm}
\paragraph{TED} \cite{ted}: An encoder-decoder transformer that is pre-trained to leverage the lead bias in millions of news articles. TED is fine-tuned on in-domain data by introducing a theme modeling objective.
\vspace{-2mm}
\paragraph{WikiTransfer} \cite{wikitransfer}: A BART-based model fine-tuned on Wikipedia-based pseudo-summaries (lead-based). For each dataset, the model is trained on pseudo-summaries that are roughly as extractive as summaries in the dataset.
\vspace{-2mm}
\paragraph{FLAN-T5} \cite{flan}: An instruction-tuned version of the T5-Large \cite{t5} encoder-decoder transformer model that has been fine-tuned on a diverse collection of NLP tasks formatted as natural language instructions.
\vspace{-2mm}
\paragraph{PROM} \cite{prom}: Augments the GSG objective with an indicator layer that identifies spans of tokens that can be copied from the source and calculates an auxiliary loss for the copying prediction.
\vspace{-2mm}
\paragraph{Centrum} \cite{centrum}: A Longformer Encoder-Decoder \cite{longformer} (LED)-based model that introduces a self-supervised pre-training objective for multi-document summarization.

We also include zero-shot performance of Mixtral Instruct \cite{mixtral} and GPT-3.5 \cite{gpt3} reported in the CCSum paper \cite{ccsum}.

For all SciTLDR summarization tasks, we use the following baselines:

\vspace{-4mm}
\paragraph{BART} \cite{bart}: An encoder-decoder transformer fine-tuned on SciTLDR article-summary pairs.
\vspace{-4mm}
\paragraph{CATTS} \citeplanguageresource{scitldr}: A BART-based model jointly fine-tuned on article-summary pairs from SciTLDR and article-title pairs (where title generation is an auxiliary task).

\vspace{-3mm}

\begin{table}
  \centering
  \small
  \begin{tabular}{l|c|c}
  \hline
   Model Name & $\alpha$ & Beam width\\
  \hline
  Llama 3.1 8B Instruct & 3 & 12\\
  Mistral Nemo Instruct 2407 & 4 & 5\\
  Gemma 2 9B It & 6 & 4\\
  \end{tabular}
  \caption{Hyper-parameters for each model are selected to maximize BARTScore \cite{bartscore} on 10\% of CNN/DM's validation split.}
    \label{alpha-vals}
\end{table}

\begin{figure}
    \centering
    \includegraphics[width=\linewidth]{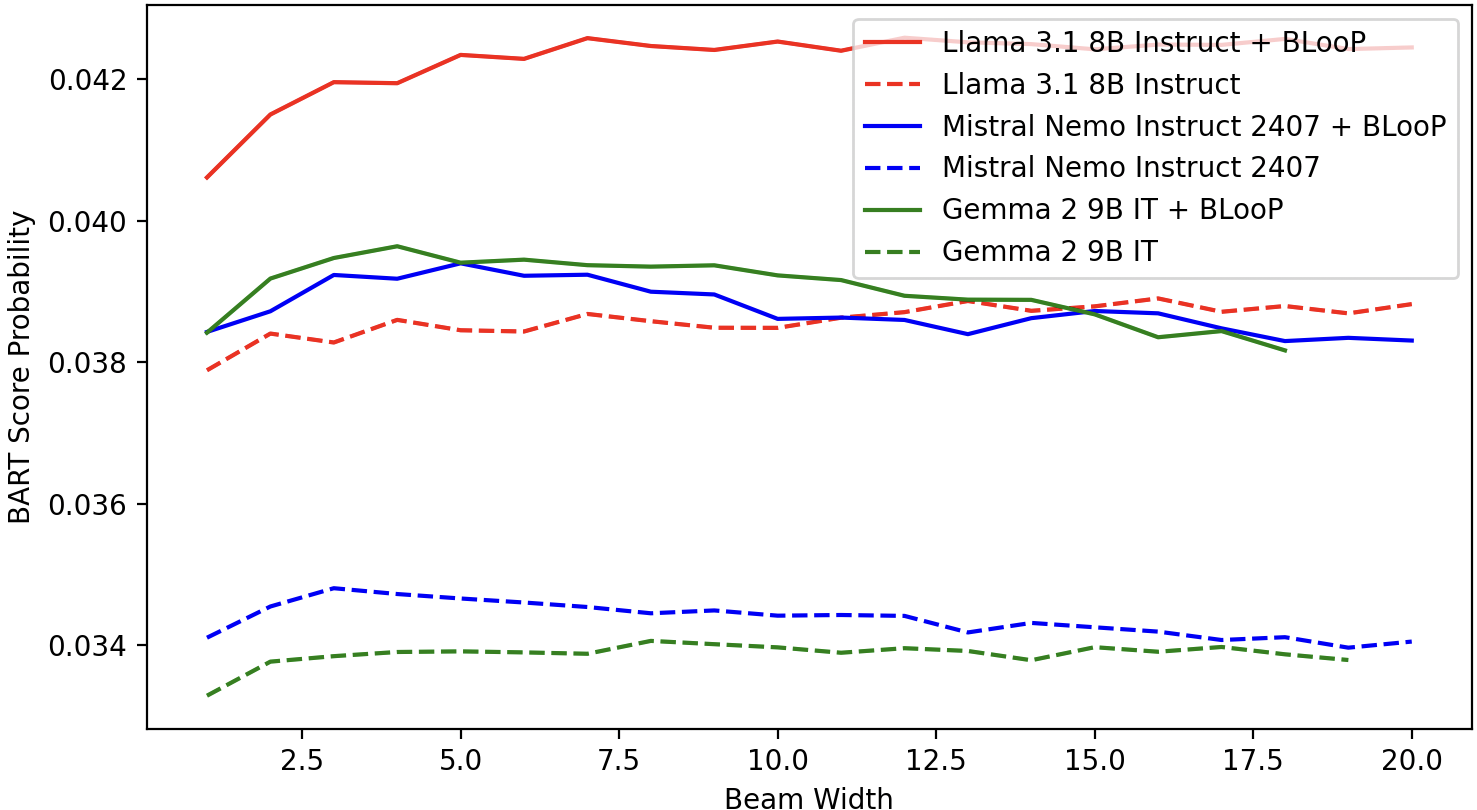}
    \caption{Increasing the beam width consistently improves Llama's performance on CNN/DM's validation split. In contrast, Gemma and Mistral stop improving once the beam width exceeds 4 and 5 respectively.}
    \label{fig:beam-perf}
\end{figure}
\paragraph{Evaluation Metrics and Hyper-parameters} Summarization performance is measured using ROUGE \cite{rouge} and BARTScore \cite{bartscore} (using the \verb|facebook/bart-large-cnn| checkpoint). We also compute a ``BARTScore Probability" metric, which is computed for each summary as $e^{\mathrm{BARTScore}}$. We evaluate similarly sized instruction-tuned \cite{it1, it2, it3} models for a fair comparison of LLMs. This means that they can use the same message format even if they each have their own chat templates. The message is formatted as shown in Listing \ref{lst:python-prompt} (where \verb|instruct| is the instruction). After forming the messages, each prompt is formed using the official chat template provided by Hugging Face for each of the three LLMs that we used in experiments, Llama-3.1-8B-Instruct, Mistral-Nemo-Instruct-2407, and Gemma-2-9b-it.

While BLooP requires no training, $\alpha$ is selected per model using grid search on 10\% of CNN/DM's validation set (maximizing BARTScore). This lightweight selection process requires only $\approx$1,100 validation examples and can be performed in a few hours with a single GPU, making deployment practical even with limited resources. Notably, the same $\alpha$ generalizes across datasets for each model (Table 1), suggesting robustness to domain shift.

The selected hyper-parameters for each model are shown in Table \ref{alpha-vals}. Temperature is set to 0 to ensure the generation is deterministic and aligns with traditional beam search. Sampling-based generation methods including top-p underperform beam search in both BARTScore and ROUGE. Figure \ref{fig:beam-perf} shows that optimal beam widths vary significantly across models. Implementation details are included in Appendix B.

In principle, $\alpha$ could be learned but BLooP aims to explore the zero-shot setting without human-annotated article-summary pairs. 

\begin{lstlisting}[language=Python, caption={Chat message used for summarization.}, label={lst:python-prompt}]
instruct = (
  "Write a paragraph summarizing the "
  "given article without preamble."
)
message = {
  "role": "user",
  "content": instruct + "\n\n" + article
}
\end{lstlisting}

\section{Results}
\subsection{Automatic Evaluation}
Table \ref{perf-table} shows that BLooP improves performance on CNN/DM, CCSum, and Multi-News. 

On CNN/DM, Llama 3.1 + BLooP achieves the best ROUGE score and Gemma 2 + BLooP achieves the best BARTScore and BARTScore Probability. Additionally, Llama 3.1 + BLooP outperforms pseudo-summary pre-trained models on CNN/DM. On CCSum, Llama 3.1 + BLooP achieves the best score on all automated metrics. Notably, BLooP enables Llama 3.1 (8B) to nearly match the zero-shot performance of GPT-3.5 (175B). On Multi-News, Gemma 2 achieves the highest performance across all metrics. This indicates that Llama 3.1 performs better when summarizing individual news articles while Gemma 2 performs better when summarizing multiple news articles. These results show that BLooP enables higher quality summarization in the news domain.

As shown in Table \ref{perf-table-scitldr}, in SciTLDR, Gemma 2 + BLooP generates the highest quality summaries as measured by both ROUGE and BARTScore metrics when the input is each paper's abstract. Moreover, even though Gemma 2 was not trained for summarization, its ROUGE-L score is only 2.9\% less than CATTS' (fine-tuned on 2k expert-annotated article-summary pairs). Its lower performance on ROUGE-1 and ROUGE-2 is significant but understandable given that the LLMs are not trained on summarization data. When the input is AIC or Full Text, BLooP underperforms in ROUGE scores compared to the abstracts-only setting. However, the best BARTScore metrics on SciTLDR are achieved when the AIC input is used with Mistral + BLooP. BLooP improves BARTScore metrics for all three LLMs on all SciTLDR input schemes. This reflects how BLooP-augmented summaries that are more semantically similar to their corresponding reference summary even when they are less lexically similar.

\begin{table}
  \centering
  \adjustbox{valign=t}{%
  \begin{scriptsize}
  \begin{tabular}{l|c|c|c|c|c}
  \hline
    Model & R-1 & R-2 & R-L & BS & BS Prob \\ \hline
  \multicolumn{6}{c}{CNN/DM}\\\hline
  Pointer-Gen $^{\dagger}$&39.53 &17.28& 36.38 &-&-\\
  PEGASUS $^{\star\dagger}$ & 44.17&21.47&41.11 & - & - \\
  PROM $^{\star\dagger}$ & 44.59& 21.66& 41.46 & - & - \\ \hline
  PEGASUS $^{\star}$ & 32.90 & 13.28 & 29.38 & - & - \\
  TED $^{\star}$ & 38.38 & 16.49 & 35.08 & - & - \\
  WikiTransfer $^{\star}$ & 39.11 &17.25 &35.73& - & - \\ 
  PROM $^{\star}$ & 37.87 &15.91& 34.16 & - & - \\ \hline
  Llama 3.1 & 40.16&16.99&36.19 &-3.44&3.73 \\
  \hspace*{\fill}+ BLooP&40.35&\underline{\textbf{18.12}}&\underline{\textbf{36.71}} &\underline{-3.39}&\underline{3.99} \\
  Mistral & 36.77&13.52&32.94&-3.43&3.72  \\
  \hspace*{\fill}+ BLooP & \underline{37.99}&\underline{16.21}&\underline{34.44}&\underline{-3.33}&\underline{4.18} \\
  Gemma 2 & 35.29&13.10&32.18&-3.43&3.74\\
  \hspace*{\fill}+ BLooP & 34.98&\underline{15.07}&32.12&\underline{\textbf{-3.32}}&\underline{\textbf{4.29}}\\\hline
  
  \multicolumn{6}{c}{CCSum}\\\hline
  FLAN-T5 $^{\dagger}$ &-&-&56.0 & - & - \\\hline
  PEGASUS $^{\star}$ & - &- &28.0 & - & - \\
  Mixtral & - &-& 38.1 & - & - \\ 
  GPT 3.5 & - &- &41.5 & - & - \\ \hline
  Llama 3.1  &46.97&28.19&39.80&-2.03&15.04 \\
  \hspace*{\fill}+ BLooP &\underline{\textbf{47.36}}&\underline{\textbf{30.28}}&\underline{\textbf{41.06}}&\underline{\textbf{-1.92}}&\underline{\textbf{16.80}} \\
  Mistral  & 37.68&18.32&30.70&-2.17&12.95 \\
  \hspace*{\fill}+ BLooP & \underline{38.74}&\underline{22.32}&\underline{32.80}&\underline{-2.01}&\underline{15.18} \\
  Gemma 2  & 34.24&16.58&28.56&-2.22&12.34 \\
  \hspace*{\fill}+ BLooP &34.30&\underline{19.38}&\underline{29.37}&\underline{-2.08}&\underline{14.39} \\\hline

  \multicolumn{6}{c}{Multi-News} \\\hline
  PEGASUS $^{\star\dagger}$ &47.52&18.72&24.91&-&-\\
  Centrum $^{\star\dagger}$ &45.7& 16.8& 23.2&-&-\\\hline
  PEGASUS $^{\star}$ & 36.54&10.52&18.67&-&-\\
  Centrum $^{\star}$ &43.5&15.7 &22.4&-&-\\\hline
  Llama 3.1 & 28.73&9.31&18.98 & -3.34 & 3.85 \\
  \hspace*{\fill}+ BLooP &\underline{29.59}&9.55&\underline{19.44}& -3.34& 3.86 \\
  Mistral  & 28.12&8.49&18.71&-3.33&3.88\\
  \hspace*{\fill}+ BLooP &\underline{29.67}&\underline{9.37}&\underline{19.47} &\underline{-3.33}&\underline{3.92} \\
  Gemma 2  & 31.02&9.08&20.68&-3.32&3.93\\
  \hspace*{\fill}+ BLooP &\underline{\textbf{31.98}}&\underline{\textbf{9.93}}&\underline{\textbf{20.85}}&\underline{\textbf{-3.31}}&\underline{\textbf{4.01}}\\\hline
  \end{tabular}
  \end{scriptsize}
  }
  \caption{Zero-shot news summarization performance of LLMs with and without BLooP. Here, R-* is ROUGE-* and BS is BARTScore. $^{\star}$ denotes pseudo-summary pre-training and $^{\dagger}$ denotes fine-tuning on the target dataset. Underlined values indicate statistically significant improvements from BLooP as measured by FDR-corrected Wilcoxon signed-rank tests ($p < 0.05$).}
  \label{perf-table}
\end{table}

\begin{table}
    \centering
    \adjustbox{valign=t}{%
    \begin{scriptsize}
    \begin{tabular}{l|c|c|c|c|c}
    \hline
      Model & R-1 & R-2 & R-L & BS & BS Prob \\ \hline 
  \multicolumn{6}{c}{SciTLDR Abstracts}\\\hline
  BART $^{\dagger}$& 43.3 & 20.8 & 35.0 & - & - \\
  CATTS $^{\dagger}$& 43.8 & 20.9& 35.5 & - & - \\ \hline
  Llama 3.1 &35.42&10.94&30.44&-4.23&1.68 \\
  \hspace*{\fill}+ BLooP &\underline{36.12}&\underline{11.58}&\underline{31.32}&\underline{-4.18}&\underline{1.78} \\
  Mistral & 35.33&9.87&30.71&-4.14&1.79\\
  \hspace*{\fill}+ BLooP & \underline{36.45}&\underline{11.41}&\underline{31.99}&\underline{\textbf{-4.08}}&\underline{1.93} \\
  Gemma 2 & 36.39&10.66&31.75&-4.11&1.88\\
  \hspace*{\fill}+ BLooP &37.05&\underline{\textbf{11.72}}&\underline{\textbf{32.65}}&\underline{\textbf{-4.08}}&\underline{\textbf{1.98}} \\\hline

  \multicolumn{6}{c}{SciTLDR AIC}\\\hline
  BART $^{\dagger}$& 42.9 & 20.8 & 35.1 & - & - \\
  CATTS $^{\dagger}$& 44.9 & 22.6& 37.3 & - & - \\ \hline
  Llama 3.1 &\textbf{36.64}&11.53&32.09&-4.10&1.93 \\
  \hspace*{\fill}+ BLooP &36.21&11.55&31.82&\underline{-4.07}&\underline{1.99} \\
  Mistral & 35.96&10.70&31.57&-4.04&1.99\\
  \hspace*{\fill}+ BLooP & 35.88&\underline{11.31}&31.54&\underline{\textbf{-4.01}}&\underline{\textbf{2.06}}\\
  Gemma 2 &36.55&11.31&\textbf{32.20} &-4.04&1.99 \\
  \hspace*{\fill}+ BLooP & 36.27&11.65&32.10&-4.03&2.02 \\\hline

  \multicolumn{6}{c}{SciTLDR Full Text}\\\hline
  Llama 3.1 &35.95&11.23&31.83&-4.09&1.98 \\
  \hspace*{\fill}+ BLooP &35.56&11.27&31.42&\underline{-4.06}&\underline{\textbf{2.04}} \\
  Mistral & 36.38&11.46&32.07&-4.08&1.94\\
  \hspace*{\fill}+ BLooP &35.91&11.68&31.59&\underline{\textbf{-4.05}}&\underline{2.01} \\
  Gemma 2 & \textbf{36.92}&11.76&\textbf{32.31} &-4.09&1.90 \\
  \hspace*{\fill}+ BLooP &36.54&11.87&32.15 &\underline{-4.06}&\underline{1.98} \\\hline
  
  \end{tabular}
  \end{scriptsize}
  }
  \caption{Zero-shot SciTLDR summarization performance of LLMs with and without BLooP.}
  \label{perf-table-scitldr}
\end{table}
\subsection{Human Evaluation}
We conducted a human evaluation to assess BLooP's impact on generated summaries' faithfulness, informativeness, and readability. Provided instructions and examples align with those used in past work \cite{prom} and are shown in Appendix C. Annotators were given summaries generated with and without BLooP for a given article and asked to determine which summary (if either) is more faithful, informative, and readable.

As shown in Table \ref{human-eval-results}, BLooP improves the faithfulness of generated summaries without significantly affecting their informativeness or readability. These results are consistent with the automatic evaluation results for news summarization tasks. Appendix D includes an article from each dataset whose summaries are improved by BLooP across all metrics and models. These examples show that BLooP enables LLMs to include more salient details from each input document.

\begin{table}
  \centering
  \small
  \begin{tabular}{l|c|c|c}
  & Win (\%) & Tie (\%) & Lose (\%) \\ \hline
  Faithfulness & 18 & 74 & 8 \\
  Informativeness & 48 & 0 & 52 \\
  Readability & 0 & 100 & 0 \\
  \end{tabular}
  \caption{Human evaluation results (\%) comparing Llama 3.1 8B Instruct with and without BLooP on 50 randomly selected examples}
  \label{human-eval-results}
\end{table}

\subsection{Abstractiveness vs. Faithfulness Trade-off}
\begin{table}[t]
  \centering
  \small
  \begin{tabular}{l|l|c|c|c}
    N & with BLooP & P (\%) & R (\%) & F1 (\%) \\\hline
    1 & No & 4.8 & 4.0 & 3.7 \\
    1 & Yes & 4.9 & 2.0 & 2.4 \\
    \hline
    2 & No & 3.2 & 3.5 & 3.0 \\
    2 & Yes & 3.6 & 1.5 & 1.8 \\
    \hline
    3 & No & 1.9 & 2.3 & 1.9 \\
    3 & Yes & 2.2 & 1.3 & 1.4 \\
  \end{tabular}
  \caption{Novel n-gram precision, recall, and F1 of Llama's predictions on the CNN/DM test set}
  \label{novel-ngram-table}
\end{table}
We accept a modest reduction in abstractiveness (as measured by novel n-gram recall) in exchange for substantial improvements in faithfulness (as measured via human evaluation and automated metrics). Three observations suggest the trade-off is favorable:
\begin{enumerate}
    \item Novel n-gram analysis (Table \ref{novel-ngram-table}) shows that BLooP improves precision of novel n-grams that appear in references, suggesting more selective rather than indiscriminate copying.
    \item Human evaluation shows no loss in readability, indicating summaries remain fluent despite increased copying.
    \item BARTScore improvements exceed ROUGE improvements across most settings, suggesting semantic quality gains beyond mere lexical overlap.
\end{enumerate}

We investigate how BLooP affects the LLMs' ability to generate novel n-grams that occur in reference summaries using a novel n-gram analysis \cite{gsum}. The analysis shown in Table \ref{novel-ngram-table} is on Llama's predictions on the CNN/DM test set.

As expected, Llama with BLooP (with a negative $\alpha$) generates fewer novel n-grams and covers fewer novel n-grams that occur in reference summaries. However, Llama with BLooP achieves slightly better novel n-gram precision across different values of n (1-3 as shown above). This indicates that BLooP pushes the model to generate a smaller selection of novel n-grams that better represent the reference summary's novel n-grams. These results suggest BLooP increases grounded reuse of source content that carries key information rather than generic copying that merely inflates lexical overlap metrics. 

\begin{figure}
    \centering
    \includegraphics[width=\linewidth]{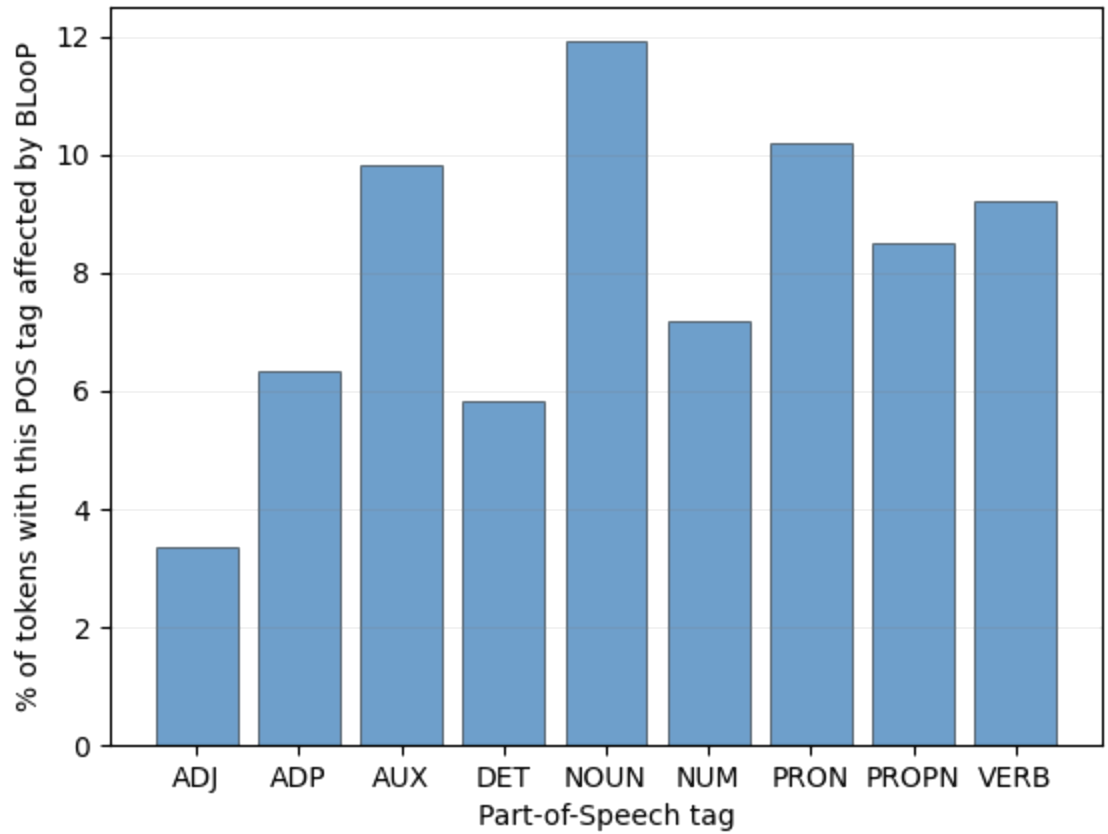}
    \caption{Part-of-speech tags of tokens in Llama-generated summaries of CCSum test set articles that differ because of BLooP.}
    \label{ccsumbar}
\end{figure}

\begin{figure*}
    \centering
    \includegraphics[width=\linewidth]{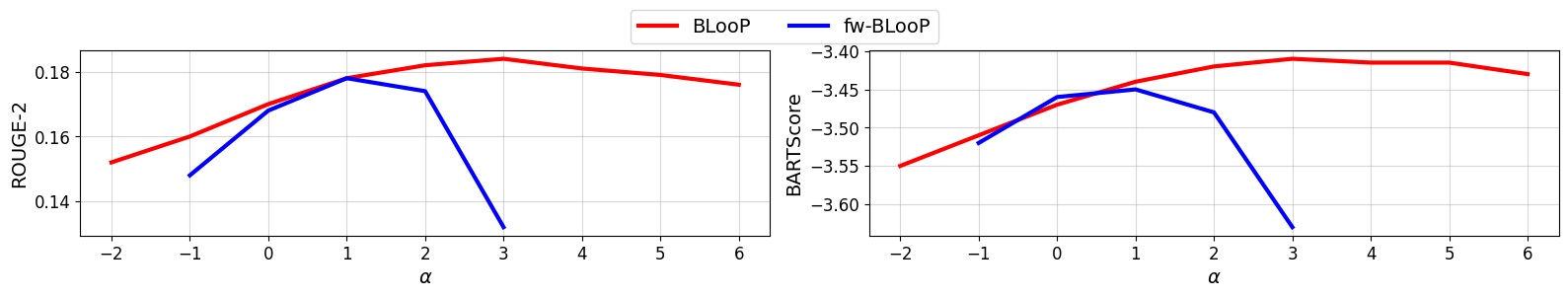}
        \vspace{-5mm}
    \caption{Ablating $\alpha$ with Llama 3.1 8B Instruct on 10\% of the CNN/DM validation set using a beam size of 8. fw-BLooP is frequency-weighted BLooP, where the promotion in Equation \ref{eq:promotion} is multiplied by the frequency of the bigram $(s_{t-1}, v)$ in the input document.}
    \label{fwcmp}
\end{figure*}
\section{Analysis}

\subsection{Part-of-Speech Tags}
We also explore the distribution of Part-of-Speech (POS) tags of tokens that changed because of BLooP. POS tags are extracted from every summary using spaCy's RoBERTa-based \cite{roberta} model $\texttt{en\_core\_web\_trf}$ \footnote{\url{https://spacy.io/models/en\#en_core_web_trf}}. Figure \ref{ccsumbar} shows that BLooP has the largest effect on generated nouns, pronouns, and (auxiliary) verbs. In contrast, BLooP has the smallest effect on predicted adjectives, adpositions, and determiners.

This suggests that BLooP primarily influences tokens that carry key information in the source document, rather than modifying descriptive or functional tokens. The heightened impact on nouns and verbs indicates that BLooP helps preserve source entities and their relationships. Furthermore, the relatively high impact on auxiliary verbs highlights BLooP's impact on the semantic structure of sentences. Examples of our model's output can be found in Appendix \ref{app2}.

\subsection{Bigram Cache Usage}
To further investigate BLooP's effectiveness, we compute \% of generation steps whose next token distribution's argmax was changed after applying the promotion in Table \ref{pen-table}. This represents how frequently BLooP changes the LLMs' predictions. On CCSum, BLooP only changes 7\% of predicted tokens but significantly improves Llama's performance across all automated metrics. This demonstrates that BLooP achieves the best performance when it can precisely change the few tokens required to steer the generation process in the right direction.

\begin{table}
  \centering
  \resizebox{0.45\textwidth}{!}{
  \begin{tabular}{|l|l|c|}
  \hline
        Dataset & Model & Argmax Change Rate \\ \hline
        \multirow{3}{*}{CNN/DM} & Gemma & 13.67 \\ 
        & Llama & 10.43 \\ 
        & Mistral & 12.60 \\ \hline
        \multirow{3}{*}{CCSum} & Gemma & 11.03 \\ 
        & Llama & 7.18 \\ 
        & Mistral & 10.08 \\ \hline
        \multirow{3}{*}{Multi-News} & Gemma & 14.69 \\ 
        & Llama & 9.87 \\ 
        & Mistral & 11.54 \\ \hline
        \multirow{3}{*}{SciTLDR-Abs} & Gemma & 13.26 \\ 
        & Llama & 9.89 \\ 
        & Mistral & 12.17 \\ \hline
        \multirow{3}{*}{SciTLDR-AIC} & Gemma & 13.19 \\ 
        & Llama & 7.17 \\ 
        & Mistral & 10.36 \\ \hline
        \multirow{3}{*}{SciTLDR-Full} & Gemma & 15.13 \\ 
        & Llama & 6.56 \\ 
        & Mistral & 7.29 \\ \hline
  \end{tabular}
  }
  \caption{\% of generation steps in which $s_t \neq s_t'$}
  \label{pen-table}
\end{table}

\begin{table}
  \centering
  \resizebox{0.39\textwidth}{!}{
  \begin{tabular}{|l|l|c|}
  \hline
Dataset & Model & Cache Hit Rate \\ \hline
        \multirow{6}{*}{CNN/DM} & Gemma & 78.36 \\ 
        & \hspace*{\fill} + BLooP & 90.46 \\ \hhline{~--}
        & Llama & 85.91 \\ 
        & \hspace*{\fill} + BLooP & 91.20 \\ \hhline{~--}
        & Mistral & 79.31 \\ 
        & \hspace*{\fill} + BLooP & 88.32 \\ \hline
        \multirow{6}{*}{CCSum} & Gemma & 81.34 \\ 
        & \hspace*{\fill} + BLooP & 92.63 \\ \hhline{~--}
        & Llama & 89.34 \\ 
        & \hspace*{\fill} + BLooP & 93.48 \\ \hhline{~--}
        & Mistral & 80.59 \\ 
        & \hspace*{\fill} + BLooP & 90.34 \\ \hline
        \multirow{6}{*}{Multi-News} & Gemma & 83.56 \\ 
        & \hspace*{\fill} + BLooP & 92.78 \\ \hhline{~--}
        & Llama & 90.61 \\ 
        & \hspace*{\fill} + BLooP & 94.39 \\ \hhline{~--}
        & Mistral & 85.80 \\ 
        & \hspace*{\fill} + BLooP & 92.88 \\ \hline
        \multirow{6}{*}{SciTLDR-Abs} & Gemma & 72.04 \\ 
        & \hspace*{\fill} + BLooP & 81.84 \\ \hhline{~--}
        & Llama & 78.03 \\ 
        & \hspace*{\fill} + BLooP & 84.20 \\ \hhline{~--}
        & Mistral & 70.72 \\ 
        & \hspace*{\fill} + BLooP & 79.51 \\ \hline
        \multirow{6}{*}{SciTLDR-AIC} & Gemma & 82.54 \\ 
        & \hspace*{\fill} + BLooP & 90.19 \\ \hhline{~--}
        & Llama & 88.95 \\ 
        & \hspace*{\fill} + BLooP & 91.95 \\ \hhline{~--}
        & Mistral & 83.43 \\ 
        & \hspace*{\fill} + BLooP & 89.76 \\ \hline
        \multirow{6}{*}{SciTLDR-Full} & Gemma & 88.04 \\ 
        & \hspace*{\fill} + BLooP & 93.82 \\ \hhline{~--}
        & Llama & 94.65 \\ 
        & \hspace*{\fill} + BLooP & 96.55 \\ \hhline{~--}
        & Mistral & 91.99 \\ 
        & \hspace*{\fill} + BLooP & 95.98 \\ \hline
  \end{tabular}
  }
  \caption{\% of bigram cache lookups in articles in which $B_{s_{t-1}} \neq \emptyset$}
  \vspace{-1mm}
  \label{cache-table}
\end{table}
Table \ref{cache-table} shows that BLooP increases cache hit rates by 5-13 percentage points, indicating increased copying from source documents. This occurs when the last predicted token is present in the source document. Most generated tokens occur in the input document regardless of whether BLooP is applied. Nonetheless, BLooP-augmented summaries contain significantly more tokens occurring in the input document. Notably, BLooP does not increase the cache hit rate in SciTLDR-AIC and SciTLDR-Full as much as in other tasks. These two tasks in particular involve long scientific documents whose bigram caches are inherently larger. Future work could analyze approaches to prune less important bigrams from the cache when summarizing long documents. 

\subsection{Frequency weighting in BLooP}
Lastly, we investigate the effect of multiplying $\alpha$ by the frequency of tokens when using BLooP. As shown in Figure \ref{fwcmp}, BLooP's performance does not improve when factoring in each bigram's frequency in the input document. This demonstrates that frequency is insufficient in determining a token's importance in articles. Future work could investigate other heuristics that can dynamically adapt the promotion in Equation \ref{eq:promotion} based on the context of a token in the article. 

\section{Conclusion}
\label{sec:conclusion}

We introduce BLooP, a novel bigram lookahead promotion designed to improve the quality of abstractive summaries generated by large language models. BLooP proactively promotes the generation of tokens that form source document bigrams. BLooP enhances the quality of generated summaries, achieving significant improvements for Llama 3.1, Mistral Nemo, and Gemma 2 in ROUGE and BARTScore across both news and scientific summarization datasets. Moreover, a human evaluation of BLooP shows that BLooP-augmented summaries are more faithful to their source document. We also provide a detailed analysis of the novel n-grams, bigram cache usage, and the prediction change rate across different summarization datasets. Our results suggest that BLooP offers a promising approach to improving summarization quality without additional pre-training or fine-tuning.

\section*{Acknowledgments}
We acknowledge grants from the National Science Foundation and Google to support this research. This work used resources available through the National Research Platform (NRP) at the University of California, San Diego. NRP has been developed, and is supported in part, by funding from National Science Foundation, from awards 1730158, 1540112, 1541349, 1826967, 2112167, 2100237, and 2120019, as well as additional funding from community partners.

\section{Bibliographical References}\label{sec:reference}

\bibliographystyle{lrec2026-natbib}
\bibliography{lrec2026-example}

\section{Language Resource References}
\label{lr:ref}
\bibliographystylelanguageresource{lrec2026-natbib}
\bibliographylanguageresource{languageresource}

\appendix

\onecolumn

\section{Statistical Significance}

\begin{table}
\centering
\small
\adjustbox{valign=t}{%
\begin{tabular}{l|c|c|c|c|c||l|c|c|c|c|c}
\hline
\multicolumn{6}{c||}{P-Values} & \multicolumn{6}{c}{Effect Sizes} \\
\hline
  Model & R-1 & R-2 & R-L & BS & BS Prob & Model & R-1 & R-2 & R-L & BS & BS Prob \\ \hline
\multicolumn{6}{c||}{CNN/DM} & \multicolumn{6}{c}{CNN/DM} \\\hline
Llama 3.1 & 1.000 & 0.000 & 0.000 & 0.000 & 0.000 & Llama 3.1 & 0.515 & 0.448 & 0.489 & 0.403 & 0.398 \\
Mistral & 0.000 & 0.000 & 0.000 & 0.000 & 0.000 & Mistral & 0.430 & 0.300 & 0.404 & 0.308 & 0.304 \\
Gemma 2 & 1.000 & 0.000 & 1.000 & 0.000 & 0.000 & Gemma 2 & 0.540 & 0.366 & 0.521 & 0.287 & 0.281 \\\hline

\multicolumn{6}{c||}{CCSum} & \multicolumn{6}{c}{CCSum} \\\hline
Llama 3.1 & 0.005 & 0.000 & 0.000 & 0.000 & 0.000 & Llama 3.1 & 0.511 & 0.430 & 0.467 & 0.315 & 0.312 \\
Mistral & 0.000 & 0.000 & 0.000 & 0.000 & 0.000 & Mistral & 0.455 & 0.280 & 0.392 & 0.252 & 0.248 \\
Gemma 2 & 1.000 & 0.000 & 0.000 & 0.000 & 0.000 & Gemma 2 & 0.520 & 0.331 & 0.466 & 0.273 & 0.270 \\\hline

\multicolumn{6}{c||}{Multi-News} & \multicolumn{6}{c}{Multi-News} \\\hline
Llama 3.1 & 0.000 & 0.256 & 0.000 & 1.000 & 1.000 & Llama 3.1 & 0.432 & 0.512 & 0.440 & 0.510 & 0.509 \\
Mistral & 0.000 & 0.000 & 0.000 & 0.044 & 0.009 & Mistral & 0.386 & 0.401 & 0.396 & 0.487 & 0.482 \\
Gemma 2 & 0.000 & 0.000 & 0.000 & 0.000 & 0.000 & Gemma 2 & 0.414 & 0.419 & 0.433 & 0.470 & 0.468 \\\hline

\multicolumn{6}{c||}{SciTLDR Abstracts} & \multicolumn{6}{c}{SciTLDR Abstracts} \\\hline
Llama 3.1 & 0.014 & 0.001 & 0.008 & 0.000 & 0.000 & Llama 3.1 & 0.462 & 0.445 & 0.451 & 0.361 & 0.366 \\
Mistral & 0.000 & 0.000 & 0.000 & 0.000 & 0.000 & Mistral & 0.410 & 0.336 & 0.391 & 0.321 & 0.325 \\
Gemma 2 & 0.089 & 0.000 & 0.028 & 0.000 & 0.000 & Gemma 2 & 0.475 & 0.384 & 0.459 & 0.408 & 0.408 \\\hline

\multicolumn{6}{c||}{SciTLDR AIC} & \multicolumn{6}{c}{SciTLDR AIC} \\\hline
Llama 3.1 & 1.000 & 1.000 & 1.000 & 0.007 & 0.006 & Llama 3.1 & 0.611 & 0.566 & 0.578 & 0.447 & 0.445 \\
Mistral & 1.000 & 0.005 & 1.000 & 0.000 & 0.000 & Mistral & 0.547 & 0.441 & 0.540 & 0.398 & 0.393 \\
Gemma 2 & 1.000 & 0.207 & 1.000 & 0.114 & 0.071 & Gemma 2 & 0.543 & 0.478 & 0.550 & 0.468 & 0.462 \\\hline

\multicolumn{6}{c||}{SciTLDR Full Text} & \multicolumn{6}{c}{SciTLDR Full Text} \\\hline
Llama 3.1 & 1.000 & 0.926 & 1.000 & 0.005 & 0.006 & Llama 3.1 & 0.607 & 0.555 & 0.570 & 0.444 & 0.447 \\
Mistral & 1.000 & 0.308 & 1.000 & 0.000 & 0.000 & Mistral & 0.591 & 0.495 & 0.562 & 0.415 & 0.406 \\
Gemma 2 & 1.000 & 0.726 & 1.000 & 0.014 & 0.012 & Gemma 2 & 0.561 & 0.505 & 0.538 & 0.445 & 0.444 \\\hline

\end{tabular}
}
\caption{Statistical significance and effect size analysis for BLooP improvements. Left: FDR-corrected p-values from Wilcoxon signed-rank tests ($p < 0.05$). Right: Effect sizes as rank-biserial correlation.}
\label{tab:significance-and-effects}
\end{table}

The FDR-corrected Wilcoxon signed-rank test results in Table \ref{tab:significance-and-effects} reveal compelling evidence for BLooP's effectiveness across diverse summarization tasks, with statistical significance achieved in 70\% of all model-dataset-metric combinations. Among these statistically significant improvements, the effect sizes range from 0.248 to 0.511 with a mean of 0.395, indicating that BLooP's significant improvements consistently fall within the medium effect range. This demonstrates that when BLooP does achieve statistical significance, the practical impact is meaningfully substantial.

\section{Implementation Details}
\label{app-imp}

We use vLLM \cite{vllm} 0.6.1 for efficient inference. When generating summaries, we use \verb|".\n"| as a \verb|stop_string| in \verb|LLM.generate()| to ensure the summaries are no longer than a single paragraph. BLooP is not applied when the argmax of the initially predicted token contains a newline or denotes the end of the sequence. This prevents BLooP from stopping a paragraph from ending (that would lengthen the summary). Summaries are generated using a traditional, deterministic beam search. Each input document is truncated such that the total prompt length for each LLM never exceeds half of the LLM's max supported context length in vLLM. This meant limiting the prompt length to 64,000 for Mistral, 4096 for Llama, and 2048 for Gemma.

The beam width and $\alpha$ hyper-parameters are selected using a grid search over beam widths from 1 to 20 and integer $\alpha$ values from -8 to 2. We use the set of hyper-parameters that maximizes BARTScore probability on 10\% of the validation split of CNN/DM.

Experiments are run inside an Ubuntu 22.04 Docker container with CUDA 12.1.0 along with the required Python dependencies. Each experiment uses 1 NVIDIA A6000 GPU, 1 x86\_64 CPU, and 35GB RAM. Hugging Face's \verb|evaluate| library is used to compute the ROUGE metrics and the original BARTScore script is used to compute the BARTScore metrics.

\newpage
\section{Human Evaluation Instructions}

The human evaluation discussed in section 5.2 was conducted using an Indian crowdsourcing company called Cogito that recruits annotators native to India who are proficient in English. We emailed their staff the instructions shown below and they consented to provide annotations. The annotators were paid \$75 for this task.

\label{heval-instruct}

\begin{small}
\begin{longtable}{|p{0.97\textwidth}|}
\hline
There will be 50 articles paired with 2 summaries each. The task is to select whether summary 1 or 2 is better in faithfulness, informativeness, and readability. The instructions are as follows:

\begin{enumerate}
    \item \textbf{Faithfulness}: Faithfulness means factual consistency with the context. Please avoid using general knowledge, and only consider it in the context of the provided document. The summary is inconsistent if facts in the summary are not supported by the document. Two typical cases are conflict and hallucination.
    \item
    \begin{enumerate}
        \item The summary contradicts the information in the document. The summary might say "A fire broke out in Seattle", but the document says it broke out in Portland. Or the summary might say "the Republicans won the election", but the document indicates the Democrats won instead.
        \item The summary adds (hallucinates) a fact that is not mentioned anywhere in the document. For example, the summary might say that "A fire broke out at 2 am", but the document does not mention the time when the fire broke out.
    \end{enumerate}
    \item \textbf{Informativeness}: It means that a summary expresses the main points of the document. A summary should contain relevant and important information and few unimportant details. If you select the summary to be not consistent with the document, please only consider the consistent information when evaluating this category.
    \item \textbf{Readability}: The summaries are written by humans or generated by language models. A summary is readable/fluent if free from language problems. A less readable summary is confusing and difficult to understand.
\end{enumerate}

We present an example below. This article reports the accidental death of Alexys Brown. The main information is the accident and the appeal to raise money and minor points can be the investigation, postmortem examination, etc. Summary 1 can be reasonable and acceptable. Summary 2 misses a major point, the appeal, thus not informative. Both summary 3 and summary 4 show unfaithfulness. Summary 3 makes a factual mistake that Alexys died of cancer. This contradicts the article. Summary 4 adds a hallucination that Alexys is three years old. Summary 5 is confusing because grammar flaws impair readability.\\\\

Article: Alexys Brown, also known as Lexi, died at her home in Emmadale Close, Weymouth, on Thursday. An investigation is under way to discover how she became trapped. A post-mortem examination is due to be carried out this week. It was originally hoped the appeal would raise £2,000. Alison Record, who started the Just Giving appeal, said she was "heart broken" over the death. “Everybody by now has heard of the terrible tragedy the Brown family have suffered with the loss of their beautiful and beloved little girl Lexi,” the appeal page reads. Many other comments have been posted on the appeal page. Steph Harris said: “Thinking of you all at this devastating time, fly high beautiful princess. Love Steph and family xxx” Lesley Andrews added: “No amount of money will take away the pain, but so much love comes with every penny. Take care. xx” Aster Group, the housing association responsible for managing the home, is assisting with the police investigation. The Health and Safety Executive (HSE) is also investigating. Dorset County Council said it had not installed the disabled lift at the property.\\\\

Summary 1:  An appeal to raise money for the family of a girl who died after getting stuck in a lift was originally hoped for raising 2,000 pounds.

Summary 2 (informativeness): Alexys Brown, also known as Lexi, died at her home in Emmadale Close, Weymouth, on Thursday.

Summary 3 (faithfulness):  Alexys Brown, also known as Lexi, died of cancer. The appeal was originally hoped for raising 2,000 pounds.

Summary 4 (faithfulness):  An appeal to raise money for Alexys Brown, a three-year-old girl who died after getting stuck in a lift was originally hoped for raising 2,000 pounds.

Summary 5 (readability):  An appeal to raise the family of Alexys Brown became trapped in a lift would raise 2,000 pounds.\\\\

When scoring summaries, 2 summaries will be provided and the task is to determine which is best (0 if they are tied):\\\\

Summary 1:  Alexys Brown, also known as Lexi, died at her home in Emmadale Close, 
Weymouth, on Thursday.

Summary 2:  Alexys Brown, also known as Lexi, died of cancer. The appeal was originally hoped for raising 2,000 pounds.\\\\

Scores:

Faithfulness: 1

Informativeness: 2

Readability: 0\\\\

The above scores denote that in faithfulness, summary 1 wins. In informativeness, summary 2 wins. In readability, the summaries are tied. The real task will include 50 news articles. On average, articles contain $\approx 650$ words and summaries contain $\approx 50$ words.\\
\hline
\end{longtable}
\end{small}

\newpage

\onecolumn
\section{Comparison of LLM-generated Summaries}
\label{app2}

\begin{small}
\begin{longtable}{|p{0.97\textwidth}|}
\hline
\textbf{CNN/DM Article}: Sheffield Wednesday chief executive Paul Aldridge will leave the Championship club at the end of next season. Aldridge, who has previously worked at West Ham, Leicester City and Manchester City, joined the club as vice chairman and CEO in January 2011. Wednesday have brought in Adam Pearson and Glenn Roeder to work alongside head coach Stuart Gray. Sheffield Wednesday chief executive Paul Aldridge (left) will leave the club at the end of next season . Chairman Dejphon Chansiri said the trio would form 'a three-man sporting director by committee' at Hillsborough. Former Hull City and Derby County owner Pearson will advise on business. Aldridge said: 'I have had a tremendous time at the club and look forward to playing my part next season in what I hope will be a very successful campaign. Glenn Roeder is now part of a new managerial committee team at the Championship outfit . Adam Pearson helped lead Hull to the Premier League and will be hoping he can do the same at Wednesday . 'I am sure by giving the chairman sufficient time to consider who is the right person to lead the club into its 150th anniversary, he will appoint the most suitable successor. 'I know I have a very busy summer ahead, we have at least two major projects starting later this month. Supporters will notice a big difference when they come back at the start of next season and see the new pitch and giant screen. 'I am also hoping to be very busy playing my part in the overhaul of the playing squad, the chairman has made it quite clear he wants to be challenging from the start of the new season.' Stuart Gray is the manager of the club and has done well since taking over from former boss Dave Jones .
\\
\hline
\textbf{Reference Summary}:  Paul Aldridge joined the club as vice chairman and CEO in January 2011 .
Aldridge had before worked at West Ham, Leieceter and Manchester City .
Glenn Roeder and Adam Pearson are part of a new management team at Sheffield Wednesday .\\
\hline
\textbf{Llama-3.1-8B-Instruct}: Sheffield Wednesday chief executive Paul Aldridge will leave the Championship club at the end of next season. Chairman Dejphon Chansiri has brought in Adam Pearson and Glenn Roeder to work alongside head coach Stuart Gray, forming a 'three-man sporting director by committee' at Hillsborough.\\
\hline
\textbf{Llama-3.1-8B-Instruct + BLooP}: Sheffield Wednesday chief executive Paul Aldridge will leave the Championship club at the end of next season. Aldridge, who has previously worked at West Ham, Leicester City, and Manchester City, joined the club as vice chairman and CEO in January 2011. Chairman Dejphon Chansiri has brought in Adam Pearson and Glenn Roeder to work alongside head coach Stuart Gray. The trio will form 'a three-man sporting director by committee' at Hillsborough.\\
\hline
\textbf{Mistral-Nemo-Instruct-2407}: 
Paul Aldridge, Sheffield Wednesday's CEO since 2011, will depart at the end of next season. The club has appointed Adam Pearson and Glenn Roeder to support head coach Stuart Gray, forming a 'three-man sporting director by committee'. Aldridge, who looks forward to a successful final season, believes chairman Dejphon Chansiri will appoint a suitable successor for the club's 150th anniversary.\\
\hline
\textbf{Mistral-Nemo-Instruct-2407 + BLooP}: Sheffield Wednesday chief executive Paul Aldridge will leave the club at the end of next season. Aldridge, who has previously worked at West Ham, Leicester City, and Manchester City, joined the club as vice chairman and CEO in January 2011. To support head coach Stuart Gray, the club has appointed Adam Pearson and Glenn Roeder to work alongside him, forming a 'three-man sporting director by committee' at Hillsborough.\\
\hline
\textbf{Gemma-2-9b-it}: Sheffield Wednesday chief executive Paul Aldridge will step down at the end of next season.  Aldridge, who joined the club in 2011, will work alongside newly appointed Adam Pearson and Glenn Roeder to form a "three-man sporting director by committee." Pearson will advise on business matters, while Roeder joins head coach Stuart Gray in the managerial team. Aldridge expressed confidence in the chairman's ability to appoint a suitable successor and highlighted upcoming projects, including a new pitch and giant screen, as well as an overhaul of the playing squad. 
\\
\hline
\textbf{Gemma-2-9b-it + BLooP}: Sheffield Wednesday chief executive Paul Aldridge will leave the club at the end of next season. Aldridge, who has previously worked at West Ham, Leicester City and Manchester City, joined the club as vice chairman and CEO in January 2011. The club has brought in Adam Pearson and Glenn Roeder to work alongside head coach Stuart Gray. Pearson will advise on business, while Roeder is now part of a new managerial committee team at the Championship outfit. Aldridge said the trio would form 'a three-man sporting director by committee' at Hillsborough. \\
\hline
\end{longtable}

\begin{longtable}{|p{0.97\textwidth}|}
\hline
\textbf{CCSum Article}: His visit is coming at a tense moment for Taiwan, China and the U.S. after House Speaker Nancy Pelosi visited Taiwan earlier this month.
TAIPEI, Taiwan -- Indiana's Republican governor met with Taiwan's President Tsai Ing-wen Monday morning, following two recent high-profile visits by U.S. politicians that drew China's ire and Chinese military drills that included firing missiles over the island.
Gov. Eric Holcomb arrived Sunday evening in Taiwan for a four-day visit that will focus on economic exchange, particularly semiconductors, according to a statement from his office.
His visit is coming at a tense moment for Taiwan, China and the U.S. after House Speaker Nancy Pelosi visited Taiwan earlier this month. China claims self-ruled Taiwan as its own territory and views exchanges with foreign governments as an infringement on its claims.
Tsai acknowledged the tensions in her opening remarks ahead of their meeting Monday morning and welcomed further exchanges.
"In the midst of this, Taiwan has been confronted by military threats from China, in and around the Taiwan Strait. At this moment, democratic allies must stand together and boost cooperation in all areas," Tsai said. "Building on our existing foundation of collaboration, I look forward to our supporting one another, and advancing hand in hand, forging closer relations and creating even deeper cooperation."
SEE MORE: Diplomatic, military tensions continue to rise in wake of Pelosi's Taiwan visit
In response to Pelosi's visit, China's military held several days of exercises that included warplanes flying toward the island and warships sailing across the midline of the Taiwan Strait, an unofficial buffer between the island and mainland.
China also imposed visa bans and other sanctions on several Taiwanese political figures, though it's unclear what effect the sanctions would have.
Holcomb emphasized the economic nature of his visit, mentioning that the state is among the top in the U.S. for direct foreign investment and was home to 10 Taiwanese companies. "We both seek to deepen and enhance our already excellent cooperation that we've established over the years," he said.
Holcomb will also meet representatives of the semiconductor industry, and is expected to promote academic and tech cooperation between Taiwan and the state of Indiana. The delegation is meeting with National Yang-Ming University and National Cheng Kung University as part of the exchange.
He is traveling with officials from the state's economic development council, as well as the dean of engineering at Purdue University, an institution that has just established a semiconductors degree program. He will visit South Korea next.
\\
\hline
\textbf{Reference Summary}: Indiana’s Republican governor began a visit to Taiwan on Monday, following two recent high-profile visits by U.S. politicians that drew China’s ire and Chinese military drills that included firing missiles over the island.\\
\hline
\textbf{Llama-3.1-8B-Instruct}: Indiana's Republican governor, Eric Holcomb, met with Taiwan's President Tsai Ing-wen to focus on economic exchange, particularly semiconductors, amidst a tense moment for Taiwan, China, and the U.S. following House Speaker Nancy Pelosi's visit to Taiwan earlier this month.\\
\hline
\textbf{Llama-3.1-8B-Instruct + BLooP}: Indiana's Republican governor, Eric Holcomb, met with Taiwan's President Tsai Ing-wen Monday morning, following two recent high-profile visits by U.S. politicians that drew China's ire and Chinese military drills that included firing missiles over the island.
The visit is focused on economic exchange, particularly semiconductors, and is coming at a tense moment for Taiwan, China, and the U.S. after House Speaker Nancy Pelosi's visit to Taiwan earlier this month.\\
\hline
\textbf{Mistral-Nemo-Instruct-2407}: Indiana's Republican Governor, Eric Holcomb, arrived in Taiwan for a four-day visit focused on economic exchange, particularly semiconductors, amidst heightened tensions between Taiwan, China, and the U.S. following House Speaker Nancy Pelosi's visit and China's subsequent military drills and sanctions.
Taiwan's President Tsai Ing-wen acknowledged the tensions and welcomed further exchanges, while Holcomb emphasized the economic nature of his visit, aiming to deepen cooperation between Indiana and Taiwanese companies.\\
\hline
\textbf{Mistral-Nemo-Instruct-2407 + BLooP}: Indiana's Republican governor, Eric Holcomb, arrived in Taiwan for a four-day visit, focusing on economic exchange, particularly semiconductors, amidst heightened tensions between Taiwan, China, and the U.S. following two recent high-profile visits by U.S. politicians that drew China's ire and Chinese military drills that included firing missiles over the island.
Taiwan's President Tsai Ing-wen acknowledged the tensions and welcomed further exchanges.
Holcomb emphasized the economic nature of his visit, mentioning that the state is among the top in the U.S. for direct foreign investment and was home to 10 Taiwanese companies.
He will also meet representatives of the semiconductor industry, and is expected to promote academic and tech cooperation between Taiwan and the state of Indiana.\\
\hline
\textbf{Gemma-2-9b-it}: Indiana's Republican governor, Eric Holcomb, met with Taiwan's President Tsai Ing-wen during a four-day visit focused on economic exchange, particularly semiconductors.
Holcomb's visit comes amid heightened tensions between Taiwan, China, and the U.S. following House Speaker Nancy Pelosi's recent trip to Taiwan, which prompted Chinese military drills and sanctions.
Tsai acknowledged the tensions and emphasized the need for democratic allies to stand together, while Holcomb stressed the economic nature of his visit, highlighting Indiana's strong ties with Taiwanese companies and its interest in deepening cooperation.
Holcomb's delegation will meet with semiconductor industry representatives and university officials to promote academic and technological exchange.
\\
\hline
\textbf{Gemma-2-9b-it + BLooP}: Indiana Governor Eric Holcomb arrived in Taiwan for a four-day visit focused on economic exchange, particularly semiconductors, following two recent high-profile visits by U.S. politicians that drew China's ire and Chinese military drills that included firing missiles over the island.
Tsai acknowledged the tensions in her opening remarks ahead of their meeting Monday morning and welcomed further exchanges.
Holcomb emphasized the economic nature of his visit, mentioning that the state is among the top in the U.S. for direct foreign investment and was home to 10 Taiwanese companies.
Holcomb will also meet representatives of the semiconductor industry, and is expected to promote academic and tech cooperation between Taiwan and the state of Indiana.
\\
\hline
\end{longtable}

\begin{longtable}{|p{0.97\textwidth}|}
\hline
\textbf{Multi-News Articles}:  LONDON (AP) — British restaurant chain Pret a Manger says a second customer has died after eating a sandwich containing an allergen that was not noted on the label. 
 
 The coffee-and-sandwich business has promised to improve its labeling following criticism at an inquest into the death of 15-year-old Natasha Ednan-Laperouse, who died in 2016 after eating a Pret baguette that contained traces of sesame. 
 
 The company said an investigation was underway into a second case, in which a customer died in December after eating a supposedly dairy-free product that contained dairy protein. Pret a Manger blamed a supplier of its dairy-free yoghurt. 
 
 The parents of Ednan-Laperouse, who are campaigning for stronger allergen warnings, said Sunday they were "incredibly saddened to learn of someone else losing their life from allergens in their food." ||||| Pret a Manger has been hit by further controversy over the weekend, as it emerged that a second customer had died after eating one of its sandwiches in 2017. \\
 
 The sandwich chain found itself in a escalating row with a former supplier, after it blamed the second death on an ingredient supplied by the vegan brand CoYo, a claim the yoghurt company said was unfounded. 
 
 It emerged on Saturday that a Pret customer had died in 2017 after eating a “super-veg rainbow flatbread” that was supposed to be dairy-free. 
 
 Pret said it had been mis-sold yoghurt that forms one of the ingredients of the flatbread and that was guaranteed dairy-free but was found to contain dairy protein. 
 
 The company agreed to full labelling of ingredients on all its freshly made products last week following the case of 15-year-old Natasha Ednan-Laperouse, who was allergic to sesame and died on a flight after eating one of its baguettes bought at Heathrow airport. 
 
 Following the news of another death, Natasha’s parents, Nadim and Tanya Ednan-Laperouse, said: “We were incredibly saddened to learn of someone else losing their life from allergens in their food. Our hearts go out to the bereaved family.” 
 
 The pair have been campaigning for allergy awareness and a change in food-labelling laws. 
 
 Pret a Manger to bring in full labelling after teenager's death Read more 
 
 The second customer, who has not been named, collapsed and died on 27 December after buying the sandwich in a shop in Stall Street, Bath. 
 
 Bath’s council alerted Pret to the incident and the chain said it withdrew all affected products. The firm said it had ended its contract with CoYo and was taking legal action. 
 
 “Subsequent testing by Pret and two independent authorities found that the CoYo dairy-free yoghurt contained traces of dairy protein,” a Pret spokesman said. “This is believed to have resulted in the tragic death of a customer from an allergic reaction in December 2017. 
 
 “Our deepest sympathies are with the family and friends of our customer in this terrible case and we will seek to assist them in any way we can.” 
 
 CoYo, a coconut milk brand that the TV cook Nigella Lawson has endorsed, said on Sunday that Pret’s claims that it was to blame were unfounded. It accused the sandwich chain of hampering its own investigation into the death by failing to provide vital information. 
 
 “Pret’s inability to provide us with a batch code, despite several requests, has severely limited our ability to investigate this further,” it said 
 
 CoYo recalled its yoghurts in February 2018 after dairy traces were found. The Food Standards Agency investigated together with Bexley council in London, where CoYo is based, before it issued an allergy alert. 
 
 CoYo denied on Sunday that the product recall was related to the death. “The dairy-free product we provided to Pret in December 2017, at the time of this tragedy, is not linked to the product we recalled in February 2018,” a spokeswoman said. 
 
 The company would continue to help to find the true cause of the death, she said. “We urge all parties to work together, and not to speculate on the cause of this tragic death that is unknown as far as we are aware and is still being investigated by the coroner’s court.” 
 
 Following the February recall, Coyo said it had been devastated to discover dairy in its yoghurts, “especially as being dairy-free sits at the heart of everything we do”. 
 
 It said it had carried out a “thorough review” of its ingredients and identified that the contaminated material was provided by an external third-party supplier, whose relationship had now been terminated.
\\
\hline
\textbf{Reference Summary}: – British restaurant chain Pret a Manger says a second customer has died after eating a sandwich containing an allergen that was not noted on the label, reports the AP. The coffee-and-sandwich business said an investigation was underway into the second case, in which a customer died December 27 after eating a supposedly dairy-free "super-veg rainbow flatbread" that contained dairy protein, reports the Guardian. "This is believed to have resulted in the tragic death of a customer from an allergic reaction," says a Pret a Manger spokesperson, and blaming CoYo, a supplier of its dairy-free yogurt. Pret a Manger has promised to improve its labeling following criticism at an inquest into the death of 15-year-old Natasha Ednan-Laperouse, who died in 2016 after eating a Pret baguette that contained traces of sesame. Natasha's parents, who are campaigning for stronger allergen warnings, said Sunday they were "incredibly saddened to learn of someone else losing their life from allergens in their food."\\
\hline
\textbf{Llama-3.1-8B-Instruct}: A second customer has died after eating a Pret a Manger sandwich that contained an allergen not noted on the label. The customer, who has not been named, collapsed and died on December 27, 2017, after eating a "super-veg rainbow flatbread" that was supposed to be dairy-free. An investigation found that the flatbread contained dairy protein, which is believed to have caused the customer's death. Pret a Manger has ended its contract with the supplier of the dairy-free yoghurt and is taking legal action.\\
\hline
\textbf{Llama-3.1-8B-Instruct + BLooP}: A second customer has died after eating a Pret a Manger sandwich containing an allergen that was not noted on the label. The customer, who has not been named, collapsed and died on December 27, 2017, after eating a supposedly dairy-free product that contained dairy protein. Pret a Manger has promised to improve its labeling following criticism at an inquest into the death of 15-year-old Natasha Ednan-Laperouse, who died in 2016 after eating a Pret baguette that contained traces of sesame.\\
\hline
\textbf{Mistral-Nemo-Instruct-2407}: Pret a Manger, a British restaurant chain, is facing another allergen-related death, the second in two years. In December 2017, a customer died after consuming a "super-veg rainbow flatbread" that was supposed to be dairy-free but contained dairy protein due to a mislabeled ingredient from supplier CoYo. Pret has ended its contract with CoYo and is taking legal action. This incident follows the death of 15-year-old Natasha Ednan-Laperouse in 2016 from an allergic reaction to sesame in a Pret baguette, leading the company to promise improved labeling.\\
\hline
\textbf{Mistral-Nemo-Instruct-2407 + BLooP}: A second customer has died after eating a Pret a Manger sandwich containing an allergen that was not noted on the label. This follows the death of 15-year-old Natasha Ednan-Laperouse, who died in 2016 after eating a Pret baguette that contained traces of sesame. Pret a Manger has promised to improve its labeling following criticism at an inquest into Ednan-Laperouse's death. The second death, in December 2017, involved a customer who died after eating a supposedly dairy-free product that contained dairy protein. Pret a Manger blamed a supplier of its dairy-free yoghurt. The parents of Ednan-Laperouse, who are campaigning for stronger allergen warnings, said they were "incredibly saddened to learn of someone else losing their life from allergens in their food."\\
\hline
\textbf{Gemma-2-9b-it}: A second customer has died after eating a Pret a Manger product containing an undeclared allergen. The customer died in December 2017 after eating a dairy-free flatbread that contained dairy protein. Pret a Manger blames a supplier, CoYo, for providing mislabeled dairy-free yoghurt. CoYo denies responsibility, stating that Pret has not provided necessary information for their investigation. Pret a Manger has ended its contract with CoYo and is taking legal action. This incident follows the 2016 death of 15-year-old Natasha Ednan-Laperouse, who died after eating a Pret baguette containing sesame, an allergen not listed on the label. 
\\
\hline
\textbf{Gemma-2-9b-it + BLooP}: A second customer has died after eating a Pret a Manger sandwich containing an allergen that was not noted on the label. The customer died in December 2017 after eating a “super-veg rainbow flatbread” that was supposed to be dairy-free but was found to contain dairy protein. Pret a Manger blamed a supplier, CoYo, a claim the yoghurt company said was unfounded. Pret a Manger has been hit by further controversy over the weekend, as it emerged that a second customer had died after eating one of its sandwiches in 2017. The sandwich chain found itself in a escalating row with a former supplier, after it blamed the second death on an ingredient supplied by the vegan brand CoYo, a claim the yoghurt company said was unfounded. Pret a Manger has promised to improve its labeling following criticism at an inquest into the death of 15-year-old Natasha Ednan-Laperouse, who died in 2016 after eating a Pret baguette that contained traces of sesame. 
\\\hline
\end{longtable}

\begin{longtable}{|p{0.97\textwidth}|}
\hline
\textbf{SciTLDR AIC}: Recent work on encoder-decoder models for sequence-to-sequence mapping has shown that integrating both temporal and spatial attentional mechanisms into neural networks increases the performance of the system substantially. We report on a new modular network architecture that applies an attentional mechanism not on temporal and spatial regions of the input, but on sensor selection for multi-sensor setups. This network called the sensor transformation attention network (STAN) is evaluated in scenarios that include the presence of natural noise or synthetic dynamic noise. We demonstrate how the attentional signal responds dynamically to changing noise levels and sensor-specific noise, leading to reduced word error rates (WERs) on both audio and visual tasks using TIDIGITS and GRID; and also on CHiME-3, a multi-microphone real-world noisy dataset. The improvement grows as more channels are corrupted as demonstrated on the CHiME-3 dataset. Moreover, the proposed STAN architecture naturally introduces a number of advantages including ease of removing sensors from existing architectures, attentional interpretability, and increased robustness to a variety of noise environments. Attentional mechanisms have shown improved performance as part of the encoder-decoder based sequence-to-sequence framework for applications such as image captioning BID22 , speech recognition BID1 BID3 , and machine translation BID0 BID21 . Dynamic and shifting attention, for example, on salient attributes within an image helps in image captioning as demonstrated by the state-of-art results on multiple benchmark datasets BID22 . Similarly, an attention-based recurrent sequence generator network can replace the Hidden Markov Model (HMM) typically used in a large vocabulary continuous speech recognition system, allowing an HMM-free RNN-based network to be trained for end-to-end speech recognition BID1 .While attentional mechanisms have mostly been applied to both spatial and temporal features, this work focuses on attention used in sensor selection. We introduce the STAN architecture that embeds an attentional mechanism for sensor selection and supports multi-sensor as well as multi-modal inputs. This attentional mechanism allows STANs to dynamically focus on sensors with higher signal-to-noise ratio (SNR) and its output is highly interpretable. Because of their inherently modular architecture, STANs remain operational even when sensors are removed after training. The same modularity makes STANs attractive for tasks that make use of multi-sensor integration. The STAN architecture can be seen as a generalization of existing less-modular network types that include attention in multi-sensor setups BID11 BID10 .This work consists of three main sections. First, we formally introduce the STAN architecture in section 2. In the first evaluation phase in section 3, we demonstrate the proper function of the attentional mechanism in synthetic noise environments with multiple audio sensors (TIDIGITS dataset) and multiple video sensors (GRID). The second evaluation phase in section 4 covers the multi-microphone real world dataset CHiME-3, where we show the proper functioning of the STAN attentional mechanism on natural noise and the robustness of STANs with respect to altered sensor configurations. The sensor\\ transformation attention network (STAN) architecture has a number of interesting features for sensor selection that we explored in this work. By equipping each sensor with an attentional mechanism for distinguishing meaningful features, networks can learn how to select, transform, and interpret the output of their sensors. Firstly, and by design, we show that STANs exhibit remarkable robustness to both real-world and synthetic dynamic noise sources. By challenging these networks during training with dynamic and persistent noise sources, the networks learn to rapidly isolate sensors corrupted by noise sources. Secondly, we show that this form of training results in even better accuracy performance from STANs than simply concatenating the sensor inputs. This is best demonstrated on the heavily noise corrupted STR environment of the CHiME-3 real-data evaluation set, where STANs achieve 23\% lower WER than concatenation models for the 50 most corrupted samples. Thirdly, we find that the output of the attention modules is highly informative, clearly indicating a sub-optimal sensor placement for a sensor pointing away from the speaker on the CHiME-3 dataset. Remarkably, this outcome is even obtained when sharing the weights of the attention modules across sensors, implying that these attention modules learned to successfully differentiate between sensors with higher and lower SNR data in presence of natural noise.Due to their modular architecture, STANs are also remarkably flexible with respect to the sensor configuration, even performing well with the removal of sensors after training. One can train STANs to solve a task with a multi-sensor setup and after training, remove the less informative sensors, with possibly savings of energy consumption and computational load on multi-sensor hardware systems with restricted computational power such as mobile robots. In the case of a defect, a sensor could be removed and STANs would remain operational with the remaining sensors.
\\
\hline
\textbf{Reference Summary}: We introduce a modular multi-sensor network architecture with an attentional mechanism that enables dynamic sensor selection on real-world noisy data from CHiME-3. A generic neural architecture able to learn the attention that must be payed to different input channels depending on the relative quality of each sensor with respect to the others.  Considers the use of attention for sensor or channel selection with results on TIDIGITS and GRID showing a benefit of attention over concatentation of features. \\
\hline
\textbf{Llama-3.1-8B-Instruct}: Researchers have developed a new neural network architecture called the sensor transformation attention network (STAN), which applies an attentional mechanism to sensor selection for multi-sensor setups. The STAN architecture is evaluated in scenarios with natural and synthetic noise, and it is shown to reduce word error rates on audio and visual tasks. The attentional signal responds dynamically to changing noise levels and sensor-specific noise, leading to improved performance. The STAN architecture also introduces advantages such as ease of removing sensors, attentional interpretability, and increased robustness to various noise environments.\\
\hline
\textbf{Llama-3.1-8B-Instruct + BLooP}: Researchers have developed a new modular network architecture called the sensor transformation attention network (STAN) that applies an attentional mechanism not on temporal and spatial regions of the input, but on sensor selection for multi-sensor setups. The STAN architecture has shown improved performance in scenarios with natural noise or synthetic dynamic noise, leading to reduced word error rates (WERs) on both audio and visual tasks using TIDIGITS and GRID; and also on CHiME-3, a multi-microphone real-world noisy dataset. The proposed STAN architecture naturally introduces a number of advantages including ease of removing sensors from existing architectures, attentional interpretability, and increased robustness to a variety of noise environments.\\
\hline
\textbf{Mistral-Nemo-Instruct-2407}: The article presents a novel neural network architecture, Sensor Transformation Attention Network (STAN), which integrates an attentional mechanism for sensor selection in multi-sensor setups. Unlike existing models that focus on temporal and spatial attention, STAN dynamically allocates attention to sensors with higher signal-to-noise ratio (SNR), improving performance in noisy environments. Evaluated on audio (TIDIGITS) and visual (GRID) datasets with synthetic noise, and a real-world multi-microphone dataset (CHiME-3) with natural noise, STAN reduced word error rates (WERs) and maintained robustness with altered sensor configurations. The modular architecture of STAN allows for easy removal of sensors post-training, offering interpretability and increased robustness to various noise environments.\\
\hline
\textbf{Mistral-Nemo-Instruct-2407 + BLooP}: The article introduces a new modular network architecture, the Sensor Transformation Attention Network (STAN), which applies an attentional mechanism not on temporal and spatial regions of the input, but on sensor selection for multi-sensor setups. Evaluated in scenarios with natural noise or synthetic dynamic noise, STAN dynamically responds to changing noise levels and sensor-specific noise, leading to reduced word error rates (WERs) on both audio and visual tasks using TIDIGITS and GRID; and also on CHiME-3, a multi-microphone real-world noisy dataset. The improvement grows as more channels are corrupted as demonstrated on the CHiME-3 dataset. The proposed STAN architecture naturally introduces advantages including ease of removing sensors from existing architectures, attentional interpretability, and increased robustness to a variety of noise environments.\\
\hline
\textbf{Gemma-2-9b-it}: The Sensor Transformation Attention Network (STAN) is a new modular network architecture that uses an attentional mechanism for sensor selection in multi-sensor setups. STAN dynamically focuses on sensors with higher signal-to-noise ratio (SNR), leading to reduced word error rates (WERs) on audio and visual tasks in both synthetic and real-world noisy environments. The attentional signal in STAN responds to changing noise levels and sensor-specific noise, improving performance as more channels are corrupted.  STAN's modular architecture allows for the removal of sensors after training, making it robust to sensor defects and adaptable to different sensor configurations. 
\\
\hline
\textbf{Gemma-2-9b-it + BLooP}: The sensor transformation attention network (STAN) is a new modular network architecture that applies an attentional mechanism for sensor selection in multi-sensor setups. STANs are evaluated in scenarios with natural noise or synthetic dynamic noise, and demonstrate reduced word error rates (WERs) on both audio and visual tasks using TIDIGITS and GRID; and also on CHiME-3, a multi-microphone real-world noisy dataset. The improvement grows as more channels are corrupted as demonstrated on the CHiME-3 dataset. STANs exhibit remarkable robustness to both real-world and synthetic dynamic noise sources, and the attention modules learned to successfully differentiate between sensors with higher and lower SNR data in presence of natural noise. Due to their modular architecture, STANs are also remarkably flexible with respect to the sensor configuration, even performing well with the removal of sensors after training. 
\\\hline
\end{longtable}

\begin{longtable}{|p{0.97\textwidth}|}
\hline
\textbf{SciTLDR AIC}: Recurrent Neural Networks (RNNs) continue to show  outstanding performance in sequence modeling tasks. However, training RNNs on long sequences often face challenges like slow inference, vanishing gradients and difficulty in capturing long term dependencies. In backpropagation through time settings, these issues are tightly coupled with the large, sequential computational graph resulting from unfolding the RNN in time. We introduce the Skip RNN model that extends existing RNN models by learning to skip state updates and shortens the effective size of the computational graph. This model can also be encouraged to perform fewer state updates through a budget constraint. We evaluate the proposed model on various tasks and show how it can reduce the number of required RNN updates while preserving, and sometimes even improving, the performance of the baseline RNN models. Source code is publicly available at https://imatge-upc.github.io/skiprnn-2017-telecombcn/. Recurrent Neural Networks (RNNs) have become the standard approach for practitioners when addressing machine learning tasks involving sequential data. Such success has been enabled by the appearance of larger datasets, more powerful computing resources and improved architectures and training algorithms. Gated units, such as the Long Short-Term Memory (Hochreiter \& Schmidhuber, 1997 ) (LSTM) and the Gated Recurrent Unit (Cho et al., 2014 ) (GRU), were designed to deal with the vanishing gradients problem commonly found in RNNs (Bengio et al., 1994) . These architectures have been popularized, in part, due to their impressive results on a variety of tasks in machine translation (Bahdanau et al., 2015) , language modeling BID17 and speech recognition (Graves et al., 2013) .Some of the main challenges of RNNs are in their training and deployment when dealing with long sequences, due to their inherently sequential behaviour. These challenges include throughput degradation, slower convergence during training and memory leakage, even for gated architectures (Neil et al., 2016) . Sequence shortening techniques, which can be seen as a sort of conditional computation (Bengio et al., 2013; Bengio, 2013; Davis \& Arel, 2013) in time, can alleviate these issues. The most common approaches, such as cropping discrete signals or reducing the sampling rate in continuous signals, are based\\ on heuristics and can be suboptimal. In contrast , we propose a model that is able to learn that samples (i.e., elements in the input sequence) need to be used in order to solve the target task. Consider a video understanding task as an example: scenes with large motion may benefit from high frame rates, whereas only a few frames are needed to capture the semantics of a mostly static scene.The main contribution of this work is a novel modification for existing RNN architectures that allows them to skip state updates, decreasing the number of sequential operations performed, without requiring any additional supervision signal. This model , called Skip RNN, adaptively determines whether the state needs to be updated or copied to the next time step. We show how the network can be encouraged to perform fewer state updates by adding a penalization term during training, allowing us to train models under different computation budgets. The proposed modification can generally be integrated with any RNN and we show, in this paper, implementations with well-known RNNs, namely LSTM and GRU. The resulting models show promising results on a series of sequence modeling tasks. In particular , we\\ evaluate the proposed Skip RNN architecture on six sequence learning problems: an adding task, sine wave frequency discrimination, digit classification, sentiment analysis in movie reviews, action classification in video, and temporal action localization in video 1 . We presented Skip RNNs as an extension to existing recurrent architectures enabling them to skip state updates thereby reducing the number of sequential operations in the computation graph. Unlike other approaches, all parameters in Skip RNN are trained with backpropagation. Experiments conducted with LSTMs and GRUs showed that Skip RNNs can match or in some cases even outperform the baseline models while relaxing their computational requirements. Skip RNNs provide faster and more stable training for long sequences and complex models, owing to gradients being backpropagated through fewer time steps resulting in a simpler optimization task. Moreover, the introduced computational savings are better suited for modern hardware than those methods that reduce the amount of computation required at each time step (Koutnik et al., 2014; Neil et al., 2016; Chung et al., 2017) .A ADDITIONAL EXPERIMENTS
\\
\hline
\textbf{Reference Summary}: A modification for existing RNN architectures that allows them to skip state updates while preserving the performance of the original architectures. Proposes the Skip RNN model that allows a recurrent network to selectively skip updating its hidden state for some inputs, leading to reduced computation at test-time. Proposes a novel RNN model where both the input and the state update of the recurrent cells are skipped adaptively for some time steps. \\
\hline
\textbf{Llama-3.1-8B-Instruct}: Recurrent Neural Networks (RNNs) have shown outstanding performance in sequence modeling tasks, but training them on long sequences can be challenging due to slow inference, vanishing gradients, and difficulty in capturing long-term dependencies. To address these issues, a new model called Skip RNN is introduced, which extends existing RNN models by learning to skip state updates and shortening the effective size of the computational graph. This model can also be encouraged to perform fewer state updates through a budget constraint, resulting in faster and more stable training for long sequences and complex models.\\
\hline
\textbf{Llama-3.1-8B-Instruct + BLooP}: Recurrent Neural Networks (RNNs) continue to show outstanding performance in sequence modeling tasks. However, training RNNs on long sequences often face challenges like slow inference, vanishing gradients and difficulty in capturing long term dependencies. To address these issues, the Skip RNN model is introduced, which extends existing RNN models by learning to skip state updates and shortens the effective size of the computational graph. This model can reduce the number of required RNN updates while preserving, and sometimes even improving, the performance of the baseline RNN models.\\
\hline
\textbf{Mistral-Nemo-Instruct-2407}: The article introduces the Skip RNN, a modification to existing Recurrent Neural Networks (RNNs) that allows them to skip state updates, reducing the number of sequential operations performed. This model adaptively decides whether to update or copy the state to the next time step, enabling it to handle long sequences more efficiently. The Skip RNN can be integrated with popular RNN architectures like LSTM and GRU, and its performance is evaluated on six sequence learning tasks. Results show that Skip RNN can match or even outperform baseline models while reducing computational requirements, leading to faster and more stable training for long sequences and complex models. The source code is publicly available.\\
\hline
\textbf{Mistral-Nemo-Instruct-2407 + BLooP}: The article introduces the Skip RNN, a novel modification for existing RNN architectures that allows them to skip state updates, decreasing the number of sequential operations performed, without requiring any additional supervision signal. This model can be integrated with any RNN, such as LSTM and GRU. The Skip RNN adaptively determines whether the state needs to be updated or copied to the next time step, and can be encouraged to perform fewer state updates by adding a penalization term during training, allowing for different computation budgets. Evaluations on six sequence learning problems show that Skip RNNs can match or even outperform baseline models while relaxing their computational requirements.\\
\hline
\textbf{Gemma-2-9b-it}: The Skip RNN model addresses the challenges of training Recurrent Neural Networks (RNNs) on long sequences by allowing the network to skip state updates. This reduces the size of the computational graph and alleviates issues like slow inference, vanishing gradients, and difficulty capturing long-term dependencies.  The Skip RNN adaptively decides whether to update or copy the state, and can be trained with a budget constraint to limit the number of updates.  Experiments with LSTM and GRU architectures demonstrate that Skip RNNs can match or improve performance on various sequence modeling tasks while requiring fewer computations, leading to faster and more stable training. 
\\
\hline
\textbf{Gemma-2-9b-it + BLooP}: The Skip RNN model addresses the challenges of RNNs, such as slow inference, vanishing gradients and difficulty in capturing long term dependencies, by learning to skip state updates and shortens the effective size of the computational graph. This model can be encouraged to perform fewer state updates through a budget constraint. Evaluations on various tasks demonstrate that Skip RNNs can reduce the number of required RNN updates while preserving, and sometimes even improving, the performance of the baseline RNN models.
\\\hline
\end{longtable}
\end{small}

\end{document}